%% file: main.tex
\def\year{2022}\relax
\documentclass[letterpaper]{article} 
\usepackage{aaai22}  
\usepackage{times}  
\usepackage{helvet}  
\usepackage{courier}  
\usepackage[hyphens]{url}  
\usepackage{graphicx} 
\usepackage{natbib}  
\usepackage{caption} 
\DeclareCaptionStyle{ruled}{labelfont=normalfont,labelsep=colon,strut=off} 
\usepackage{algorithm}
\usepackage{algorithmic}

\usepackage{newfloat}
\usepackage{listings}
\floatstyle{ruled}
\newfloat{listing}{tb}{lst}{}
\floatname{listing}{Listing}

\usepackage{amsmath,amsfonts,amssymb}
\usepackage{caption}
\usepackage{xcolor}
\usepackage{subfig}
\DeclareMathOperator*{\argmax}{arg\,max}

\DeclareMathOperator{\Unif}{Unif}

\DeclareMathOperator{\p}{\mathbb{P}}

\newcommand{\floor}[1]{\left\lfloor#1\right\rfloor}

\usepackage{amsthm}
\newtheorem{problem}{Problem}

\usepackage{tikz}
\usetikzlibrary{shapes.geometric, arrows}
\usetikzlibrary{positioning,fit,calc,arrows.meta}
\tikzstyle{startstop} = [rectangle, minimum width=0.5cm, minimum height=0.2cm,text centered, draw=black, fill=red!30]
\tikzstyle{arrow} = [thick,->,>=stealth]

\title{AlphaSnake: Policy Iteration on a Nondeterministic NP-hard Markov Decision Process}
\author {
    Kevin Du\textsuperscript{\rm 1},
    Ian Gemp\textsuperscript{\rm 2},
    Yi Wu\textsuperscript{\rm 3},
    Yingying Wu\textsuperscript{\rm 4,5,\thanks{corresponding author}}
}
\affiliations {
    \textsuperscript{\rm 1} Harvard University, Cambridge, U.S. \\
    \textsuperscript{\rm 2} DeepMind, London, U.K.\\
    \textsuperscript{\rm 3} Institute for Interdisciplinary Information Sciences, Tsinghua University, Beijing, China\\
    \textsuperscript{\rm 4}  Center of Mathematical Sciences and Applications, Harvard University, Cambridge, U.S.\\
    \textsuperscript{\rm 5} University of Houston, Department of Mathematics, Houston, U.S.\\
    kevindu@college.harvard.edu, imgemp@deepmind.com,
    jxwuyi@mail.tsinghua.edu.cn,
    ywu68@uh.edu
}

\begin{document}

\maketitle

\begin{abstract}

Reinforcement learning has been used to approach well-known NP-hard combinatorial problems in graph theory. Among these, Hamiltonian cycle problems are exceptionally difficult to analyze, even when restricted to individual instances of structurally complex graphs. In this paper, we use Monte Carlo Tree Search (MCTS), the search algorithm behind many state-of-the-art reinforcement learning algorithms such as AlphaZero, to create autonomous agents that learn to play the game of Snake, a game centered on properties of Hamiltonian cycles on grid graphs. The game of Snake can be formulated as a single-player discounted Markov Decision Process (MDP) where the agent must behave optimally in a stochastic environment. Determining the optimal policy for Snake, defined as the policy that maximizes the probability of winning \textemdash \, or win rate \textemdash \, with higher priority and minimizes the expected number of time steps to win with lower priority, is conjectured to be NP-hard. Performance-wise, compared to prior work in the Snake game, our algorithm is the first to achieve a win rate over $0.5$ (a uniform random policy achieves a win rate $< 2.57 \times 10^{-15}$), demonstrating the versatility of AlphaZero in tackling NP-hard problems.
\nocite{GraphAGo,Hamiltonian,AlphaZero}

\end{abstract}

\subsection{Introduction}

General-purpose reinforcement learning algorithms have made significant progress in achieving superhuman performance in games \cite{AlphaZero}. Recent work has investigated the potential of applying these algorithms to complexity theory \textemdash \, in particular, to solve NP-hard graph problems for which there are no known polynomial time solutions. A common framework for formulating these problems is the deterministic MDP \cite{GraphAGo}, where an agent optimizes its performance in a deterministic environment.


In this paper, however, we formulate the game of Snake as a nondeterministic environment in which the agent must maximize its expected rewards, in resemblance to the inherent uncertainty in real-world problems. Prior work has used Deep Q-Learning and tree-based Monte Carlo models to train autonomous agents in Snake, but our work is the first to use AlphaZero's policy iteration and search algorithms. Performance-wise, although it is evaluated in a relatively smaller environment, our algorithm appears to be the first to achieve a win rate over $0.5$, demonstrating the versatility of AlphaZero in addressing NP-hard problems.

\subsection{The Snake Environment}

The game of Snake is played on an $n \times n$ board, where the player controls the movement of the snake head, moving one unit in three of the four orthogonal directions every time step. The snake body follows the head's movements and grows by one unit when the snake eats an apple, increasing the score of the player by one. Once the apple is eaten, another apple is placed in an empty cell uniformly at random. The snake head can not intersect the snake body, and if the head is surrounded by the body, the game is lost; this game rule forces the path defined by the snake body to be a Hamiltonian path. Starting from a snake size of 2 units, the goal of the game is to maximize the length of the snake. Minimizing the win time, or number of time steps required to achieve the maximum snake size, is a secondary objective. Figure 1 shows various example game states.

\begin{figure}[!h]
    \centering
    \includegraphics[width=2cm]{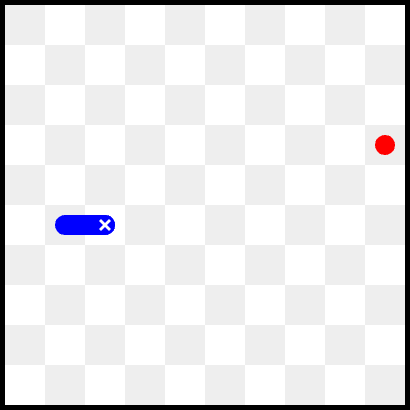}
    \includegraphics[width=2cm]{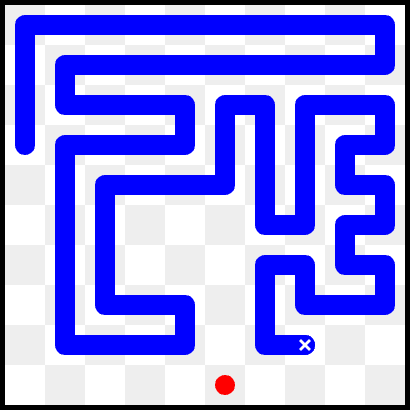}
    \includegraphics[width=2cm]{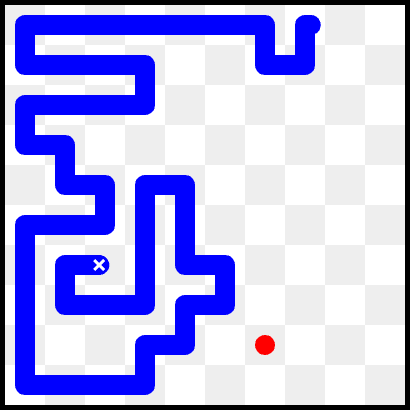}
    \includegraphics[width=2cm]{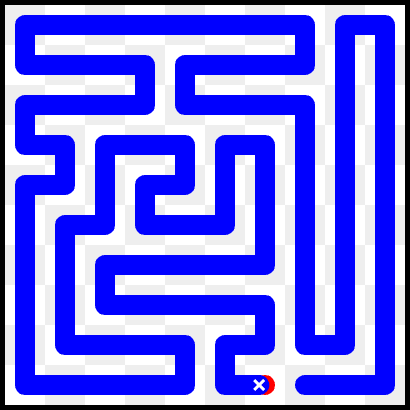}
    \caption{Game states from left to right: example (1) starting state (2) intermediate state (3) losing state (4) winning state.}
    \label{fig:my_label}
\end{figure}

Due to stochasticity in the environment, Snake can be formulated as a nondeterministic MDP, in which finding the optimal policy requires the agent to maximize its expected rewards. The rewards used in this project are $+1$ for eating an apple, $-10$ for a losing end state, and $+10$ for a winning end state. These rewards were chosen arbitrarily, but we found experimentally that this scheme was effective.

\subsection{Deterministic Algorithms and Complexity}

\begin{table}[!h]
    \centering
    \begin{tabular}{ c | c | c }
     Strategy & Average score & Win rate \\ 
     \hline
     AlphaZero algorithm & 98.227 & 944/1000 \\  
     Uniform random policy & 6.277 & 0/1000 \\
     Hamiltonian cycle strategy & 30.026 & 0/1000 \\
     Naive tree search & 59.09 & 0/100\\
    \end{tabular}
    \caption{Average score and win rate, as fraction of wins within 1,200 steps.}
    \label{tab:my_label}
\end{table}

There exists a trivial algorithm that is guaranteed to win Snake with a worst-case win time of $\mathcal{O}(n^4)$ (Supp. Eqn.~2). This is to simply follow a Hamiltonian cycle that traverses the $n \times n$ grid graph, which guarantees that the snake will not intersect itself and will eat any apple placed in the grid. However, this strategy does not minimize the expected win time, and is thus not optimal.

The optimal policy for Snake is defined as the policy that, for an arbitrary game state, maximizes the win rate with higher priority and minimizes the expected win time with lower priority. It is known that the optimal policy for Snake has an expected win time of at least $\mathcal{O}(n^3)$ (Supp. Eqn.~3). It is also conjectured that determining the optimal policy for Snake is NP-hard, since the problem of extending an arbitrary partial path on a grid graph to a Hamiltonian cycle can be reduced to finding the optimal policy for Snake (Supp. Problem 1). To better approximate this optimal policy, we adopt the algorithm behind AlphaZero.

\subsection{Adapting AlphaZero to Discounted MDPs}

To modify the AlphaZero algorithm to the game of Snake, a few differences between Snake and AlphaZero's target games \textemdash \, chess, Go, and Shogi \textemdash \, are identified:

\paragraph{Low action count:} Snake has far fewer actions than Chess or Go, with at most three legal moves in any position. In fact, many states in Snake have only one possible action. Thus, in our implementation of the Snake environment, we remove these single-action states by simulating forced actions.
\paragraph{Long horizon:} While the games of Chess and Go typically last under 300 moves, the game of Snake requires $\mathcal{O}(n^3)$ time steps. Furthermore, intermediate rewards in Snake are easier to identify, since a higher score in intermediate game states is a strong signal of progress. Thus, we use a discounted MDP to model Snake.


To adapt the AlphaZero algorithm to MDPs, we will need a value network that gives, instead of the expected final outcome of the game as in AlphaZero, the discounted value:
\[
V(s) = \mathbb{E}[R(s) + \gamma^{t(s')-t(s)} R(s') + \gamma^{t(s'')-t(s)} R(s'') + \cdots]
\]
\noindent
where $s, s', s'', \ldots$ are the future states of the game, $R(s)$ is the reward given in the state $s$, and $t(s)$ is the time index in state $s$. Just like in AlphaZero, MCTS with leaf evaluation given by the value network is used to compute the improved policy $\pi$, which is used to simulate rollouts. However, instead of two players who alternate between minimizing and maximizing the expected value, a single player seeks to maximize the value while a ``chance'' agent branches into new nodes uniformly at random to simulate stochasticity. 


\begin{figure}[!h]
    \centering
    \includegraphics[width=4.15cm]{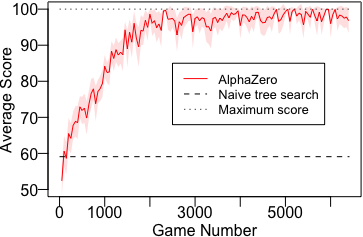}
    \includegraphics[width=4.15cm]{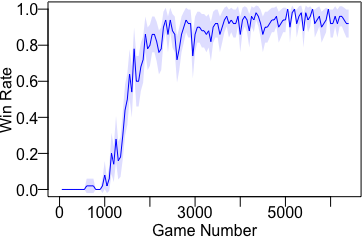}
    \caption{(Left) Graph of scores in comparison to Naive Tree Search, (Right) Graph of win rate averaged every 50 games. Shaded areas are $2\sigma$ error bars.}
    \label{fig:my_label}
\end{figure}


The value network is trained with mean-squared loss on the estimates of discounted return obtained from rollouts.
The policy network is trained with cross-entropy towards the state-action counts of tree search (see Supp. for details).



\subsection{Experiments}

In our experiments, we ran the above algorithm on a $10 \times 10$ game, using a tree search size of 200 states for 6,000 games.

Table 1 compares this performance with several other algorithms for Snake: a random policy, the Hamiltonian cycle strategy, and a naive tree search (without policy or value predictions) with a search size of 10,000 states.




\normalfont

\subsection{Conclusion}

Our work shows that AlphaZero is resilient in performing in a nondeterministic game, developing strategies that are successful irrespective of chance. The game of Snake may be a relevant model for real-world security systems and anti-poaching strategies that often require protecting targets from adversary response \cite{Security}. Generalizing AlphaZero to stochastic multiagent environments appears to be promising for tackling real-world situations.

\subsection{Acknowledgments}

We would like to thank Clifford Taubes and Richard Peng for discussions on computational complexity.



\bibliography{aaai22,supp}


\appendix
\section{Supplementary}
\input{supp_arxiv.tex}

\end{document}

%% file: supp_arxiv.tex
\subsection{AlphaZero and Monte Carlo Tree Search}

AlphaZero has achieved superhuman performance in a variety of challenging games such as chess and Go \cite{AlphaZero}. In this paper, we modify the AlphaZero algorithm to analyze the game of Snake, a nondeterministic environment with a longer time horizon.


The two basic components of AlphaZero are the convolutional neural network giving policy and value predictions and the MCTS algorithm generating rollouts of the game.

The neural network $f_\theta$ with parameters $\theta$ takes as input a game state $s$ and outputs a policy vector $\mathbf{p}$ and a value $v$. The vector $\mathbf{p}(a)$ is a vector of probabilities for selecting each move $a$ in state $s$ and the value $v$ represents the expected final outcome of the game. In AlphaZero, the rewards were originally set to $+1$ for a win, $-1$ for a loss, and $0$ for a draw.

To simulate a rollout, moves are selected sequentially for each player using MCTS to explore the search space. During the tree search, leaf nodes are added to the search tree in sequence, determined by the Predictor-Based Upper Confidence Trees (PUCT) algorithm which selects nodes that have the maximum PUCT value \cite{AlphaGoZero}:
\[
a^* = \argmax_a \left[ Q(a) + c_{puct} \mathbf{p}(a) \frac{\sqrt{\sum_b N(b)}}{1 + N(a)} \right]
\]
\noindent
where $Q(a)$ is the current average value of action $a$, $\mathbf{p}(a)$ is the policy given by the network, and $N(a)$ is the current visit count of the node represented by action $a$. Once a leaf node is reached, it is evaluated by the network and the search tree is updated. After the tree search exploration, move probabilities are calculated as $\pi(a) \sim N(a)^{1/\tau}$, where $\tau$ is a temperature parameter.

After the rollout is completed, the network values of $(\mathbf{p}, v)$ are trained towards $(\pi, z)$, where $\pi$ is the probability vector with each entry indicating the probability of selecting each move direction during the rollout and $z$ is the final outcome. The loss function for this training is
\[
l = (v - z)^2 + \pi^{\intercal} \log(\mathbf{p}) + c ||\theta||,
\]
where $c$ is the parameter to control L2 regularization~\cite{AlphaGoZero}.

\subsection{AlphaZero on Nondeterministic Environments}

AlphaZero's algorithm used MCTS in deterministic settings, but our implementation modifies MCTS to account for stochastic transitions. The stochasticity in Snake is solely due to the uniform random placement of the next apple after the snake consumes the current apple. To modify MCTS, each action the snake can take is represented as an intermediate node. If the action is stochastic, the corresponding node branches into states after the random event, with each state representing one possible new location of the apple. Since the event of the apple appearing in each empty cell is equally likely, our algorithm explores each branch with equal frequency.

\subsection{Complexity of Snake}

It is conjectured that the game of Snake is NP-hard, as there exists a reduction from the Partial Path Problem to the Snake game:

\begin{problem}[Partial Path Problem]
Given an arbitrary path $P$ on $G_n$, the $n \times n$ grid graph, does there exist a Hamiltonian cycle $H$ on $G_n$ that contains $P$?
\end{problem}

This is because any path in $G_n$ can be represented as a snake in the game, and the winning sequence of moves is equivalent to completing a Hamiltonian cycle. It is known that the problem of determining whether there is a Hamiltonian cycle on an arbitrary grid graph is NP-hard \cite{hamilton}. Thus, the Snake game is NP-hard if the board shape is allowed to be arbitrary \cite{snakeHard}, but the case of $n \times n$ boards is yet to be analyzed.

\subsection{Deterministic Algorithms for Snake}

The deterministic Hamiltonian Cycle strategy is guaranteed to win Snake, where the win time follows the distribution $T = \sum_{i=1}^{n^2-2} X_i$ where $X_i \sim \Unif(\{1, \ldots, i\})$ are independent. This is because the number of time steps to reach the apple when there are $i$ units of remaining space is uniformly distributed from $1,\ldots,i$ depending on the location of the apple. In the case of a $10 \times 10$ board, by the central limit theorem, the probability of winning under the time limit of $1,200$ time steps is roughly,
%
%
\begin{equation}
\p[T \leq 1,200] \approx 2.566 \times 10^{-15}
\end{equation}

Note that a uniform random strategy has a strictly smaller win rate, as the expected number of time steps to reach a randomly selected cell is strictly larger.

\subsection{Asymptotic Bounds of The Hamiltonian Cycle Strategy}
The worst-case of win time for the Hamiltonian cycle strategy is:
\begin{equation}
\max(T) = 1 + 2 + \ldots (n^2-2) = \Theta(n^4),
\end{equation}
because the worst-case number of time steps to reach the next apple when the snake has size $k$ is $n^2-k$, while the snake starts with size 2. 
To be precise, denote the Hamiltonian cycle as
$\{v_1, v_2, \ldots, v_{n^2}\}$, where $v_i$ signifies the location of the $i$th element on the Hamiltonian cycle on the board. For the sake of concreteness, let the initial state of the snake have the snake head at $v_1$ and the body at $v_2$. In the worst case scenario, the first apple will appear at $v_3$, which takes the snake $n^2 - 2$ steps to eat by moving along the sequence $v_{n^2}, v_{n^2-1}, \ldots, v_3$. After the snake eats the apple at $v_3$, the head is displaced at $v_3$, and the body is located at $v_4$ and $v_5$. The second apple will appear at $v_6$, which takes the snake $n^2 - 3$ steps to eat with the head moving along the sequence $v_2, v_1, v_{n^2}, v_{n^2-1}, \ldots, v_6$. In summary, it takes the snake $n^2-2$ steps to eat Apple 1, $n^2-3$ steps to eat Apple 2, and so on until 1 step to eat Apple $n^2 - 2$.
This process is deterministic, so the asymptotic bound is tight.

\subsection{The Lower Bound of The Optimal Strategy}

Consider a square of side length $2\floor{n\over 4} - 1$ centered at the head of the snake. If the apple is not within this square, it will be at a distance of at least $\left\lfloor{n \over 4}\right\rfloor$ from the head of the snake.
The number of such cells is at least
$$n^2 - \left(2\floor{n\over 4} - 1\right)^2
\geq n^2 - \left(2\times{n\over 4}\right)^2 = {3\over 4}n^2.$$
For the first ${n^2\over 2}-2$ apples, the length of the snake is at most ${n^2\over 2}$.
Since ${n^2\over 2} < {3\over 4}n^2$, there is always a square of distance at least $\floor{n\over 4}$ from the head. Therefore, to reach the first ${n^2\over 2} - 2$ apples, the snake travels at least distance 
$$
\left({n^2\over 2}-2\right) \cdot \floor{n\over 4} > 
\left({n^2\over 2}-2\right) \cdot \left({n\over 4}-1\right)
=\Omega(n^3).
$$


\subsection{Network Architecture}

The input to the neural network $f_\theta$ is a stack of seven $10 \times 10$ binary feature planes. Four planes contain binary values representing the presence of a snake unit pointing in each of four directions. The last three planes contain all zero entries, except for a single entry that represents the positions of the head, tail, and apple.

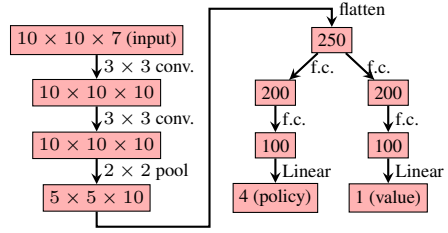
\begin{figure}[!h]
    \centering
    {\scriptsize
    \begin{tikzpicture}[node distance=0.7cm, level distance=0.7cm, sibling distance = 1.5cm]
    \node (start) [startstop] {$10 \times 10 \times 7$ (input)};
    \node (n1) [startstop, below of = start] {$10 \times 10 \times 10$};
    \node (n2) [startstop, below of = n1] {$10 \times 10 \times 10$};
    \node (n3) [startstop, below of = n2] {$5 \times 5 \times 10$};
    \node (n4) [startstop, right=1.7cm of start] {250}
        child {node (p1) [startstop] {200}}
        child {node (v1) [startstop] {200}};
    \node (p2) [startstop, below of = p1] {100};
    \node (p3) [startstop, below of = p2] {4 (policy)};
    \node (v2) [startstop, below of = v1] {100};
    \node (v3) [startstop, below of = v2] {1 (value)};
    \draw [arrow] (start) -- node[anchor=west] {$3 \times 3$ conv.} (n1);
    \draw [arrow] (n1) -- node[anchor=west] {$3 \times 3$ conv.} (n2);
    \draw [arrow] (n2) -- node[anchor=west] {$2 \times 2$ pool} (n3);
    \draw [arrow] (n3.south) -| ++(0,-0.2) -| ++(1.5,2.9) -| node[anchor=west] {flatten} (n4.north);
    \draw [arrow] (n4) -- node[anchor=west] {f.c.} (p1);
    \draw [arrow] (p1) -- node[anchor=west] {f.c.} (p2);
    \draw [arrow] (p2) -- node[anchor=west] {Linear} (p3);
    \draw [arrow] (n4) -- node[anchor=west] {f.c.} (v1);
    \draw [arrow] (v1) -- node[anchor=west] {f.c.} (v2);
    \draw [arrow] (v2) -- node[anchor=west] {Linear} (v3);
    \end{tikzpicture}
    }
    \caption{Summary of the network architecture. Convolutional layers have $3 \times 3$ kernel, stride 1, and padding 1; $2 \times 2$ downsampling layers use Max pooling; f.c. represents fully connected layers; and leaky ReLu nonlinearity is used, except on arrows labelled ``Linear'' which are fully connected layers with linear activation.
}
\end{figure}

The output of the network is a vector of four logit probabilities representing each of the four orthogonal directions in which the snake can move, as well as a scalar representing the value.

\subsection{Hyperparameters}

The learning rate used in this project was $0.001$ and the momentum was $0.7$. The temperature for determining $\pi$ from the visit counts of MCTS was $\tau = 0.5$. The exploration constant of the PUCT algorithm was $c_{puct} = 0.5$.

Training games were simulated in a $10 \times 10$ snake board with a time limit of $1,200$ time steps. Training consists of mini-batches of size $100$ states from the last $2000$ games, with $30$ batches trained after every game. The discount factor used was $\gamma = 0.98$.

\subsection{Comparison in Performance with Prior Work}

Deep Q-Learning is frequently applied in the game of Snake, utilizing different training mechanisms such as dual experience replay \cite{snakeWei}, tree-based Monte Carlo models \cite{snakeWu}, state space compression \cite{snakeSebastianelli}, and Meta-Gradients \cite{snakeBonnet}. A comparison of our algorithm with these Deep Q-Learning approaches is seen in Table 2.

\begin{table}[!h]
\fontsize{8}{12}\selectfont
\centering
\begin{tabular}{ c | c | c }
 Algorithm & Average Score & Board Size \\ 
 \hline
 AlphaZero Algorithm & 98\% & $10 \times 10$ \\  
 Dual Experience Replay & $8\%^{\text{\cite{snakeWei}}}$ & $12 \times 12$ \\
 State Space Compression & $14\%^{\text{\cite{snakeSebastianelli}}}$ & $20 \times 20$ \\
 Tree-Based Model & $19\%^{\text{\cite{snakeWu}}}$ & $12 \times 12$ \\
 Meta-Gradients & $80\%^{\text{\cite{snakeBonnet}}}$ & $12 \times 12$
\end{tabular}
\caption{A comparison of average score and board size across multiple training mechanisms. To standardize ``average score'' across different implementations of the Snake game, we calculate the ratio of average snake size in the terminal state to the area of the board. Thus, achieving a maximum score of $n^2$ on an $n \times n$ board would result in a score of $100\%$. The scores are reported from the cited papers.}
\end{table}

\subsection{Learned Strategies}

Over the course of 6000 games, the agent appears to learn many strategies to improve in the game. The strategies measured in our experiments are described in Figure 4 and displayed in Figure 5.

\onecolumn

\begin{figure}[!h]
    \centering
    \includegraphics[width=0.8\textwidth]{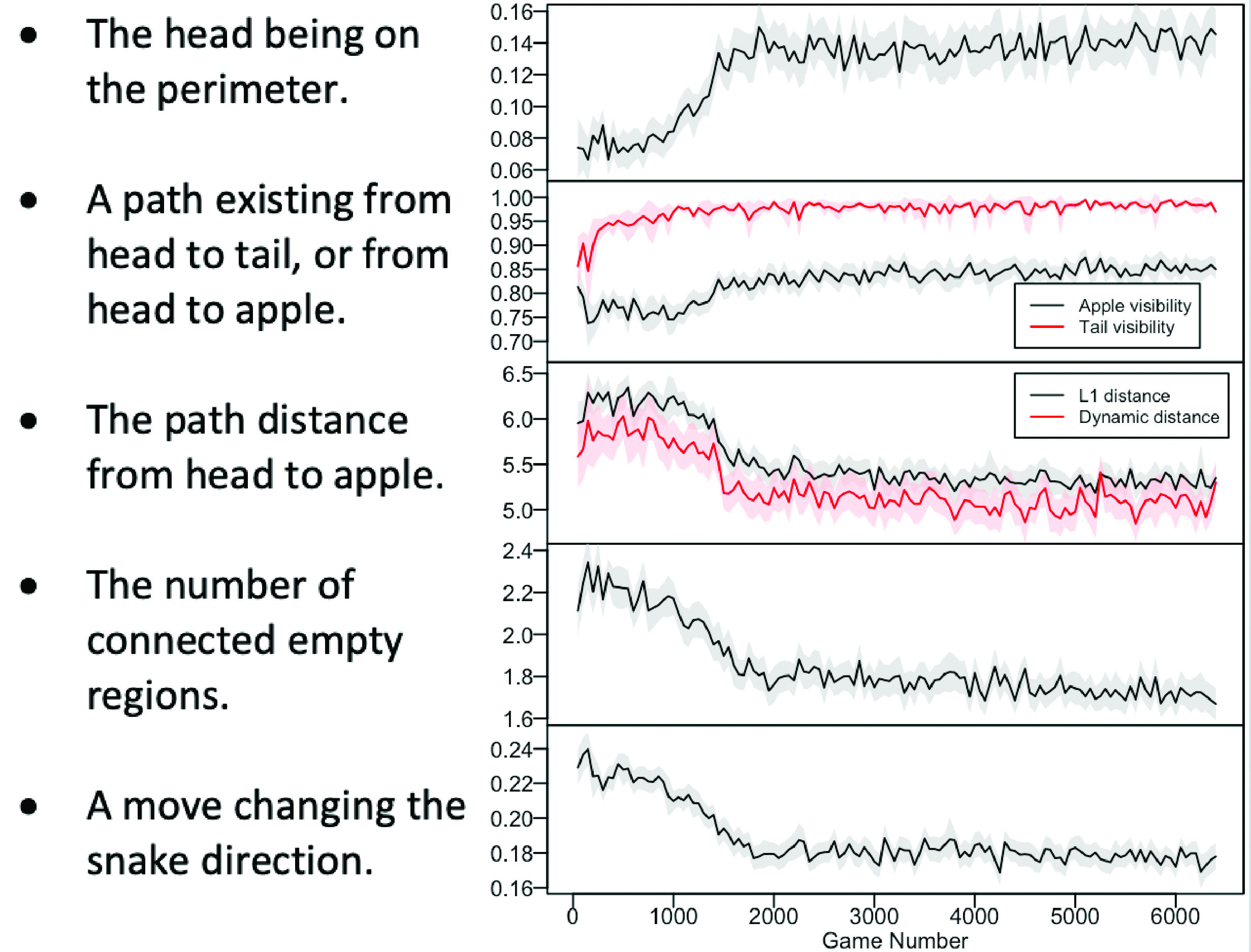}
    \caption{Developments of certain strategies over the course of training \footnotemark.}
    \label{fig:my_label}
\end{figure}










\begin{figure}[!h]
    \centering
    \includegraphics[width=0.2\textwidth]{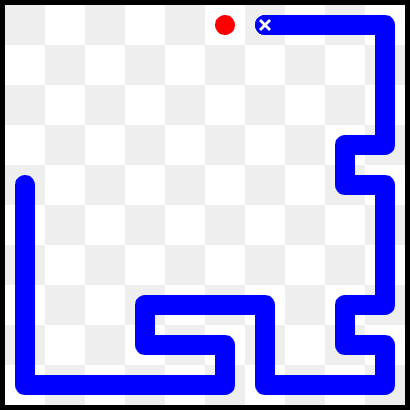}
    \includegraphics[width=0.2\textwidth]{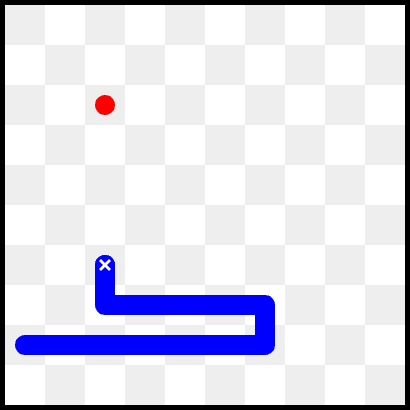}
    \includegraphics[width=0.2\textwidth]{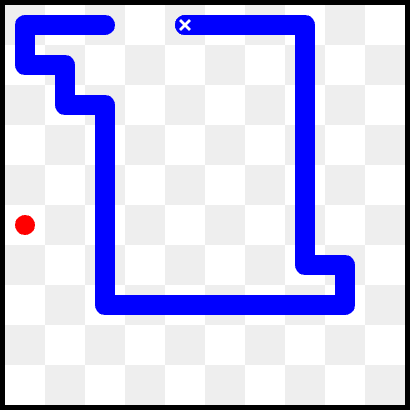}
    \includegraphics[width=0.2\textwidth]{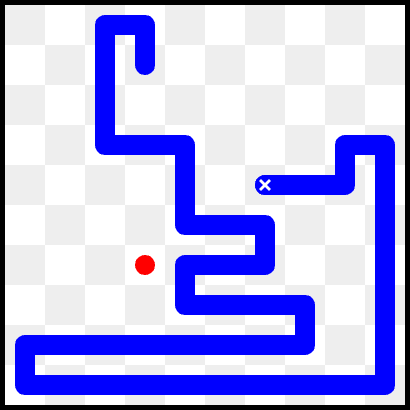}
    \caption{Game states from left to right: example states where (1) the head is on the perimeter, (2) the apple is visible from the head, (3) the apple is not visible from the head, (4) there are three connected empty regions - one on the left, one on the top right, and one on the center right. }
    \label{fig:my_label}
    \vspace{-5mm}
\end{figure}


\footnotetext{Dynamic distance refers to the number of time steps the snake must take to reach the apple if it always moves along the shortest path to the apple that does not intersect the snake body.}


